\documentclass[sigconf]{acmart}

\usepackage{booktabs} 
\usepackage{listings}
\usepackage{xcolor}
\usepackage{caption}
\usepackage{subcaption}
\usepackage{multirow}
\usepackage{siunitx}
\usepackage{natbib}

\definecolor{codegreen}{rgb}{0,0.6,0}
\definecolor{codegray}{rgb}{0.5,0.5,0.5}
\definecolor{codered}{rgb}{1,0,0}
\definecolor{stringgray}{rgb}{0.4,0.4,0.4} 
\definecolor{backcolour}{rgb}{1,1,1} 

\lstdefinestyle{mystyle}{
  backgroundcolor=\color{backcolour},   
  commentstyle=\color{codegreen},       
  keywordstyle=\color{codered},         
  numberstyle=\tiny\color{codegray},    
  stringstyle=\color{stringgray},     
  basicstyle=\ttfamily\footnotesize,    
  breakatwhitespace=false,         
  breaklines=true,                 
  captionpos=b,                    
  keepspaces=true,                 
  numbers=left,                    
  numbersep=5pt,                  
  showspaces=false,                
  showstringspaces=false,
  showtabs=false,                  
  tabsize=2
}

\setcopyright{rightsretained}

\renewcommand\footnotetextcopyrightpermission[1]{}

\acmDOI{}

\acmISBN{}

\renewcommand\footnotetextcopyrightpermission[1]{} 
\acmYear{2025}
\copyrightyear{2025}

\acmArticle{}
\acmPrice{}

\editor{}
\editor{}
\editor{}

\begin{document}
\title{Semantic Decomposition and Selective Context Filtering}
\subtitle{Text Processing Techniques for Context-Aware NLP-Based Systems}

\author{Karl John Villardar}
\affiliation{%
  \institution{Cebu Institute of Technology}
  \streetaddress{Natalio B. Bacalso Ave}
  \city{Cebu City}
  \country{Philippines}
  \postcode{6000}
}
\email{karljohn.villardar@cit.edu}

\begin{abstract}
In this paper, we present two techniques for use in context-aware systems: \textit{Semantic Decomposition}, which sequentially decomposes input prompts into a structured and hierarchal information schema in which systems can parse and process easily, and \textit{Selective Context Filtering}, which enables systems to systematically filter out specific irrelevant sections of contextual information that is fed through a system's NLP-based pipeline. We will explore how context-aware systems and applications can utilize these two techniques in order to implement dynamic LLM-to-system interfaces, improve an LLM's ability to generate more contextually cohesive user-facing responses, and optimize complex automated workflows and pipelines.
\end{abstract}

%
%



\maketitle

\section{Introduction}

Recent advances in Large Language Models (LLMs) have led to them becoming increasingly more capable and versatile via various scaling laws related to model parameter count and training dataset size \cite{kaplanScalingLawsNeural2020, minaeeLargeLanguageModels2024}. These developments have significantly improved the models' abilities to understand context, generate coherent responses, and adapt to a wide range of tasks without extensive retraining. As a result, LLMs are now better equipped to mimic human-like language comprehension and generation, enabling them to perform complex analytical tasks, such as summarizing lengthy documents, generating creative content, and providing insightful predictions \cite{minaeeLargeLanguageModels2024}. Furthermore, their ability to learn and generalize from diverse datasets has opened new avenues for innovative applications, driving further research and deployment in various sectors.

Consequently, there is a growing need for effectively capturing the nuances of natural language inputs and transforming them into formats that systems can easily parse and process \cite{bucaioniFunctionalSoftwareReference2025}. This is because natural language is notoriously difficult to work with and design systems around due to it's free-form and fluid nature. For domains that deal with large amounts of complex natural language data such as Customer Service, Healthcare, Education, Finance, \cite{urlanaLLMsIndustrialLens2024, bommasaniOpportunitiesRisksFoundation2022}, or any industry that requires constant processing of new documents \cite{bucaioniFunctionalSoftwareReference2025}, LLMs are becoming more desirable for integration into automated systems within these domains to enhance their decision-making capabilities, streamline redundant processes, and reduce operational costs in some specific sub-areas of these domains \cite{urlanaLLMsIndustrialLens2024}.

In order to enable a more cohesive integration of LLMs and implement a more scalable and a better structured natural language data processing schema for these systems, we propose two new techniques: \textit{Semantic Decomposition} and \textit{Selective Context Filtering}. These techniques induces a form of synthetic, controlled, and human-interpretable "systematic reasoning" within these systems, and can be integrated with other existing LLM-based techniques as well to improve an LLM-based agent's text generation to be more coherent with respect to its integrated system's context.

\subsection{Context-Aware Systems}

We define \textit{Context-Aware Systems} as those that leverage LLM-based pipelines to dynamically adapt and respond to a wide range of contextual inputs. These systems are designed to understand, incorporate, and react to various forms of input data, allowing them to perform tasks with a high degree of relevance and precision. Integrating context-awareness in systems using LLM-based pipelines enhances their functionality, adaptability, and user-centricity across varied application domains.  

One such example can be found in customer service platforms, where LLMs analyze customer inquiries and provide real-time assistance by drawing on context from prior interactions, customer history, and specific queries to offer personalized responses \cite{meinckeReimaginingCustomerService, bommasaniOpportunitiesRisksFoundation2022}.

In the healthcare domain, context-aware systems powered by LLMs can synthesize patient data and medical records to assist healthcare professionals in diagnosing conditions or recommending treatment plans. By utilizing specific patient histories and current symptomatology, these systems ensure that their outputs are tailored to the unique needs of each patient, thereby enhancing decision support processes \cite{munnangiAssessingLimitationsLarge2024}.

Educational technologies also benefit from context-aware systems, where LLMs personalize learning experiences by adapting to a student's progress and comprehension level. Such systems take into account a student's previous interactions, learning pace, and areas of difficulty to curate educational content and suggest resources that best suit the student's current understanding, promoting an individualized learning journey \cite{bommasaniOpportunitiesRisksFoundation2022, liAdaptingLargeLanguage2024}.

In the realm of finance, context-aware systems utilize LLMs to analyze market trends, economic data, and investor behavior, providing financial analysts with insights and forecasts that are highly contextualized. By integrating real-time data streams and historical market information, these systems support timely and informed investment decisions \cite{liLargeLanguageModels2024}.

In summary, context-aware systems utilize the power of LLM-based pipelines to enhance their functionality, adaptability, and user-centricity across varied application domains. They enable systems to interact more intelligently with users by continuously refining and applying contextual information, leading to more accurate outcomes and improved user experiences. As these systems improve over time, they promise to transform industries by offering scalable, personalized, and dynamically adaptive solutions, thereby meeting the complex demands of modern applications and significantly improving decision-making processes.

\begin{figure*}[t]
    \centering
    \begin{subfigure}{0.48\textwidth} 
        \centering
        \includegraphics[width=\textwidth]{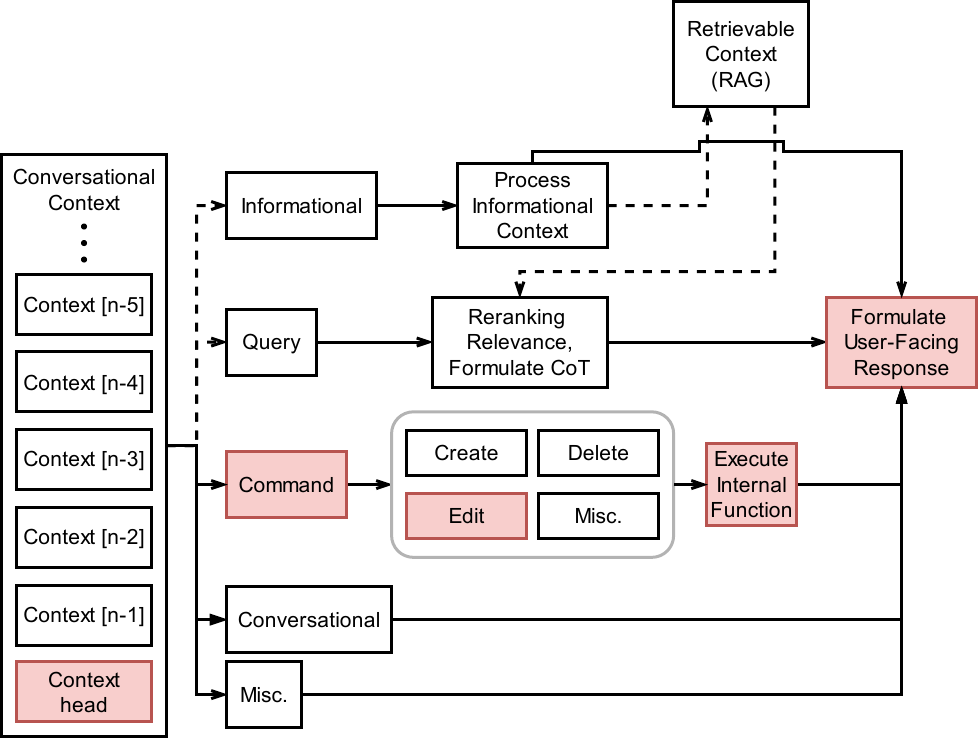}
        \caption{Command}
        \label{fig:command}
    \end{subfigure}
    \hfill
    \begin{subfigure}{0.48\textwidth} 
        \centering
        \includegraphics[width=\textwidth]{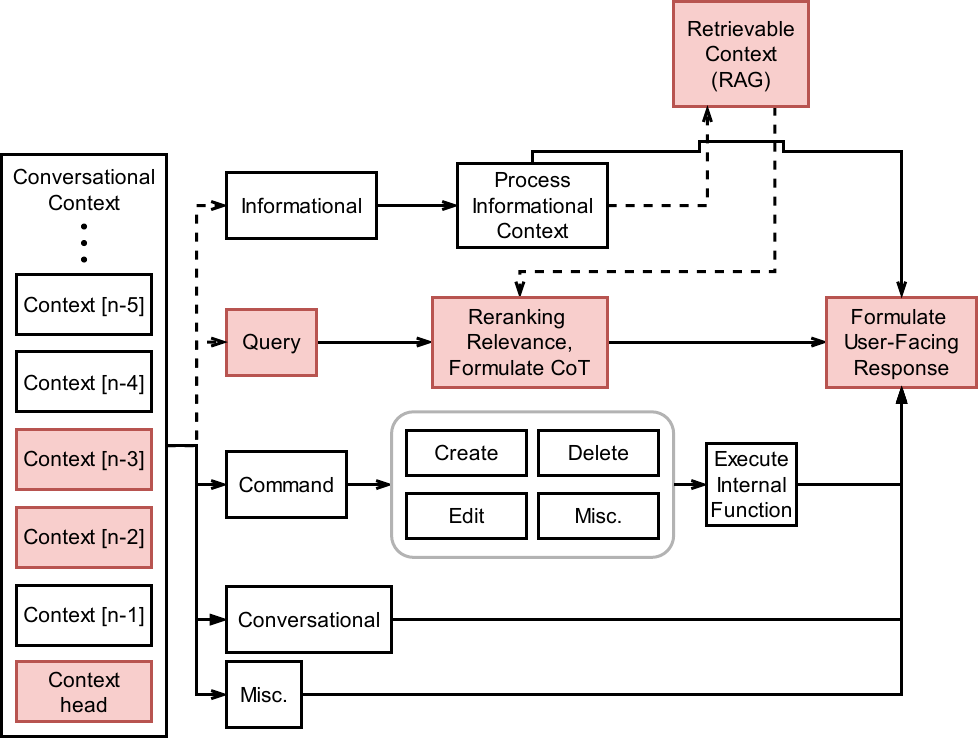}
        \caption{Query}
        \label{fig:query}
    \end{subfigure}
    \caption{Example system schema where \textit{Semantic Decomposition} and \textit{Selective Context Filtering} is implemented. Highlighted in red shows the relevant, filtered-down context and the pipeline path that is used to generate the user-facing prompt.}
    \label{fig:system}
\end{figure*}

\section{Limitations of Large Language Models}
\subsection{Context Window Size}

Context in LLMs is the set of specific input data that is used to generate an output response. More specifically, it is the set of text tokens that is fed into an LLM's transformer network, which combined with the transformer's attention mechanism, generates another set of output tokens that is a plausibly valid and contextually coherent sequence continuation of the set of input tokens. This is because the attention mechanism within transformers takes "important" tokens within the current context and assigns high relevance values for them \cite{vaswaniAttentionAllYou2023}, so your output sequence will be coherent with respect to your input.

However, the transformer attention mechanism has a very inherent limitation, and that is its limited context window size \cite{gongSelfAttentionLimitsWorking2024}. This limitation comes from the fact that the attention mechanism requires a quadratic \((O(n^2))\) scaling in computational requirements as the input sequence length grows \cite{vaswaniAttentionAllYou2023}. This effectively means that the longer your LLM generates an output sequence, the more likely that the head of the sequence will lose coherence with respect to the original input context.

Recent advances have explored various methods to extend the context window by altering the attention mechanism to be more efficient in order to improve the LLM's ability to handle longer sequences \cite{wangLimitsSurveyTechniques2024, almanFundamentalLimitationsSubquadratic2024}. However, system designers and engineers aiming to integrate these LLMs within their systems still need to work around this context window size limitation because most use-cases can easily surpass the context size of even the largest transformer models in the current market\footnote{Within the current market as of late 2024 \cite{minaeeLargeLanguageModels2024}.}. 

\subsection{Representational Memory Limits}

Representational memory refers to the model's capability to retain and utilize information across different contexts and tasks. LLMs primarily rely on their weights to store and retrieve information. These weights encode vast amounts of knowledge during training. For an LLM, the ability to store unique information as memory is dictated by the effective parameter size of the model, meanwhile the ability to generate unique output responses to inputs is dictated by the size of the training dataset distribution \cite{kaplanScalingLawsNeural2020}.

However, while LLMs have shown remarkable prowess in capturing short-term dependencies, they often struggle with long-term information retention and retrieval, which is crucial for tasks requiring sustained reasoning or recalling information presented earlier in a dialogue or document. Moreover, during inference, the model lacks explicit mechanisms to access or update this knowledge dynamically based on new contexts. As a result, LLMs may produce outputs that seem disconnected from prior established facts or exhibit inconsistency across an extended output sequences.

\subsection{Hallucination}

Hallucination in LLMs is a phenomenon wherein the generated output sequence is plausible, coherent, and grammatically correct, but is factually incorrect or logically nonsensical at times. The previously mentioned limitations largely contribute to hallucination phenomenon. Output sequences becoming incoherent with respect to the original context the longer it grows, or the model producing outputs that seem inconsistent from earlier priors are all forms of hallucination \cite{huangSurveyHallucinationLarge2025}. 

Despite the representational complexity within LLMs, they are ultimately autoregressive next-token prediction models \cite{vaswaniAttentionAllYou2023}. As long as the output sequence is probabilistically plausible according to the LLM's training data distribution, there will be a likelihood that hallucination will occur. Therefore, external methodologies are necessary in order for our LLM agents to produce coherent output consistently. 

\section{Existing Techniques for Enhancing LLM Outputs}
\subsection{Retrieval Augmented Generation}

Retrieval Augmented Generation (RAG) enhances the capabilities of Large Language Models (LLMs) by integrating information retrieval mechanisms into the text generation framework \cite{lewisRetrievalAugmentedGenerationKnowledgeIntensive2021}. This approach is particularly effective in improving the accuracy and relevance of the output by accessing and utilizing external information from pre-existing corpora or knowledge bases, which may not be directly encoded in the LLM's training data \cite{gaoRetrievalAugmentedGenerationLarge2024}.

\subsubsection{Vector Embeddings}
RAG is typically implemented using vector embeddings, where both the query (input context or prompt) and the documents in the knowledge base are transformed into fixed-size vector representations \cite{lewisRetrievalAugmentedGenerationKnowledgeIntensive2021}. These vector embeddings are generated using pre-trained models, such as BERT or other transformer-based architectures, which capture semantic similarities between texts \cite{lewisRetrievalAugmentedGenerationKnowledgeIntensive2021, gaoRetrievalAugmentedGenerationLarge2024}. When a query is made, the system compares the vector embedding of the query with those of the documents to retrieve the most relevant information. This retrieved information is then combined with the original input to generate more contextually aware responses.

\subsubsection{Graph Embeddings}
In addition to vector embeddings, graph embeddings are also utilized to implement RAG. Graph embeddings involve representing entities and their relationships in a knowledge graph, where nodes represent concepts or entities, and edges denote the relationships between them \cite{velickovicGraphAttentionNetworks2018}. By embedding these graph structures into continuous vector spaces, LLMs can retrieve and integrate relational information that is structured and interconnected. This method is beneficial for applications that require a deep understanding of relationships and context, such as question answering or dialogue systems \cite{edgeLocalGlobalGraph2024}.

Both vector and graph embedding approaches enable RAG to dynamically access external information and incorporate it into the language generation process, resulting in enhanced outputs that are more informative and relevant.

\subsection{Chain-of-Thought}

Chain-of-Thought (CoT) is a technique used to emulate human-like reasoning patterns. It involves breaking down complex problems or tasks into a series of manageable, sequential steps, allowing models to follow a logical progression in their response generation \cite{weiChainofThoughtPromptingElicits2023}.

In the context of LLMs, implementing CoT involves training models to generate intermediate reasoning steps before arriving at a final conclusion or response. This approach helps the model to maintain coherence and logical consistency, as each step builds on the previous one. Chain-of-Thought is particularly valuable in tasks that require deductive reasoning, problem-solving, and multi-turn dialogue management, where the system must track and integrate various pieces of information over several interactions \cite{weiChainofThoughtPromptingElicits2023, huangReasoningLargeLanguage2023}.

To leverage CoT effectively, models can be designed to output intermediate steps as part of their response generation, encouraging transparency in the decision-making process. Additionally, during training, incorporating datasets that model logical reasoning paths and employing techniques such as reinforcement learning can enhance the model's ability to execute a coherent and structured chain of reasoning \cite{weiChainofThoughtPromptingElicits2023}. As a result, CoT aids in improving the interpretability and reliability of LLM outputs, making them more suitable for complex and nuanced applications \cite{huangReasoningLargeLanguage2023}.

\subsection{Structured Outputs}

Structured Outputs refer to the capability of Large Language Models (LLMs) to generate responses that adhere to a predefined format or structure, rather than producing free-form text \cite{willardEfficientGuidedGeneration2023}. This approach is particularly beneficial in applications that require precise and consistent information presentation, such as generating tables, filling out forms, producing JSON objects, or creating code snippets with specific syntax \cite{shortenStructuredRAGJSONResponse2024}.

To implement structured outputs, LLMs are typically trained with data that includes examples of the desired output format. This training helps the model understand the rules and structure necessary to produce outputs that conform to specific templates. Moreover, prompt engineering can be used effectively to guide the model towards generating structured data. By providing well-formatted examples and specifying clear instructions in the input prompt, users can influence the LLM to follow a particular structure in its responses \cite{shortenStructuredRAGJSONResponse2024, liuWeNeedStructured2024}.

Another key technique in achieving structured outputs involves aligning the model's language generation process with task-specific constraints. For instance, using schemas or templates as a guide during generation can help LLMs maintain consistency in data fields and attributes. Additionally, incorporating post-processing validation steps ensures that the generated outputs meet structural requirements, correcting any deviations or errors introduced during the generation phase \cite{shortenStructuredRAGJSONResponse2024}.

But the most important method for ensuring the consistency of structured outputs is the utilization of Context-Free Grammars (CFGs) \cite{willardEfficientGuidedGeneration2023}. CFGs provide a formal way to define the permissible structure of output by specifying the syntax rules that the generated text must adhere to. This method is particularly valuable in enforcing uniformity across outputs, as CFGs can systematically constrain the model’s outputs to match the desired structural patterns. By integrating CFGs into the generation process, LLMs can be nudged to produce outputs that are not only semantically rich but also structurally consistent with the specified requirements.

Structured Outputs are essential in domains where uniformity and clarity are crucial, such as data entry automation, report generation, and interactions with APIs. By ensuring that outputs are structured, LLMs can improve accuracy, facilitate downstream processing, and provide more user-friendly and actionable information \cite{liuWeNeedStructured2024}. As models evolve, expanding their capabilities to produce various and complex structured outputs will remain a pivotal area of research and application development.

\section{Approach}

In this section, we propose two novel techniques designed to enhance the performance and adaptability of context-aware systems: \textit{Semantic Decomposition} and \textit{Selective Context Filtering}. These methodologies aim to improve how information is structured and utilized within these systems, enabling them to interact with Large Language Models (LLMs) more effectively.

\subsection{Semantic Decomposition}

This technique involves sequentially breaking down input prompts based on a predefined, hierarchal manner. The aim is to let an LLM agent pipeline break down input prompts in a way that conforms to a system's required schema. By decomposing prompts into discrete, manageable segments, automated systems can interpret and execute complex pipelines based on natural language commands with precision, thereby facilitating more accurate and relevant response generation.

Unlike in CoT, this process not only optimizes the system's input parsing capabilities, but this also improves the fine-grain control of the system's "reasoning process". This is because the reasoning of the system is bounded within the schema's framework that the system designer supplies the LLM agents with. 

\begin{figure}[h]
\centering
\includegraphics[width=0.45\textwidth]{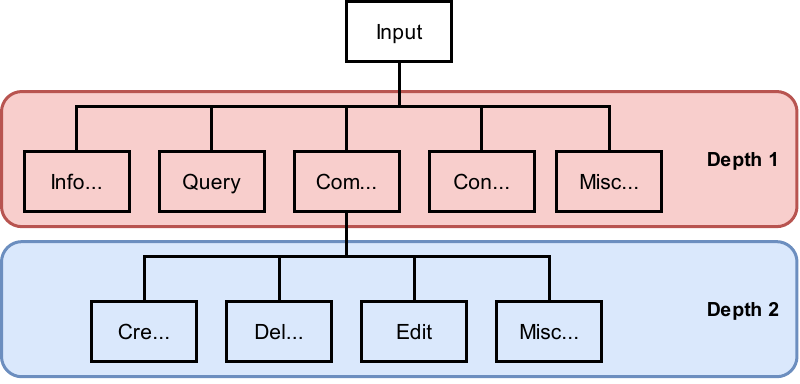}
\caption{Example of a two-level Semantic Decomposition schema for a single processing pipeline. Inputs prompts can either be \textit{Informational}, a \textit{Query}, \textit{Conversational},or a \textit{Command}. A \textit{Command} can either be \textit{Create}, \textit{Delete}, or \textit{Edit}. We use \textit{Miscellaneous} as our negative fall-back category. }
\end{figure}

The process involves instructing our LLMs to discern and categorize the inputs from a set of pre-defined classes from a schema. We sequentially go down the schema's depth if the discerned category if the category is not a leaf class of the hierarchy. We essentially turn our LLM agent into a pseudo-classification model via Structured Outputs, so our output is guaranteed to follow a schema that is parsable by the integrated system.

\begin{lstlisting}[language=Python, frame=single, caption=Schema Implementation of Depth 1 (from Figure 2) using Python.]
from pydantic import BaseModel, Field
from typing import Optional
from enum import Enum

class InputType(str, Enum):
    """
    The input type of the prompt as a whole.
    - query: The user's intent is a query type. The user is asking a question or wishes to acquire information.
    - command: The user's intent is a command. The user is asking you, the system to perform a task.
    - informative: The user's intent is informational. This means that the user is trying to inform you, the system, a piece of context or information.
    - conversational: The user is greeting the system, or there is no intrinsic subject or topic within the user's prompt. 
    - miscellaneous: If the user is prompting nonsensical or incoherent inputs.
    """
    query = 'query'
    command = 'command'
    informative = 'informative'
    conversational = 'conversational'
    miscellaneous = 'miscellaneous'

class InputProcessingSchema(BaseModel):
    input_type: InputType = Field(None, alias="Input Type")
    subject: Optional[str] = Field(
        title='Subject',
        description='The core subject matter, the focus, or main semantic idea of the prompt. This is the area of interest that the user has provided.'
    )
    context_instructions: Optional[str] = Field(
        title='Context Instructions',
        description='The context in which the subject matter is encapsulated in. This includes what the user has instructed to you, the system.'
    )
\end{lstlisting}

Along each level of decomposition, we can choose to re-summarize our input prompt in a more concise manner that fits the current context, or we can just pass the original prompt data along the pipeline. The former tends to result in better categorization, as we mimic the logical refinement found in CoT due to the re-summarization context being more semantically dense and concise, containing only the important information relevant to the passed context. 

\begin{lstlisting}[frame=single, caption={Semantic Decomposition of a command prompt, along with decomposed context instructions.}]
Input: "Create a new document named 'Exercises'. Have it under folder 'Examples'. Set the contents to be blank for now."
Output: {
    "input_type": "command",
    "subject": "Document Creation",
    "context_instructions": "Create a new document named 'Exercises' in the 'Examples' folder with blank content."
}
\end{lstlisting}

Note that our enumeration definition for the input type is guaranteed by the LLM due to CFGs in Structured Outputs. And it is very important to give our type enumerations \textit{negative fall-back} categories. This is important if we want our pipelines to have well defined behavior. In our given example, we used "miscellaneous" as our negative 
fall-back category. 

\begin{lstlisting}[frame=single, caption=Semantic Decomposition of a nonsensical prompt.]
Input: "asdkWldqwlqwlej."
Output: {
    "input_type": "miscellaneous",
    "subject": null,
    "context_instructions": null
}
\end{lstlisting}

\subsection{Selective Context Filtering}

Current LLM-based text generation pipelines often involve inputting whole sections of context into an LLM (usually the last nth items of an input context), along with external retrieved context snippets (via RAG), without refinement. It is a well understood property of LLMs that irrelevant snippets of information in your inputs may lead to short-term decoherence. This is why most modern RAG algorithms perform re-ranking in order to alleviate decoherence due to uncorrelated pieces of retrieved context.

To approach this issue, we propose \textit{Selective Context Filtering} in which we embed our sections of input context, and consecutively remove irrelevant sections with embedding-based filters as the context data moves along our pipeline. 

\begin{figure}[h]
\centering
\includegraphics[width=0.45\textwidth]{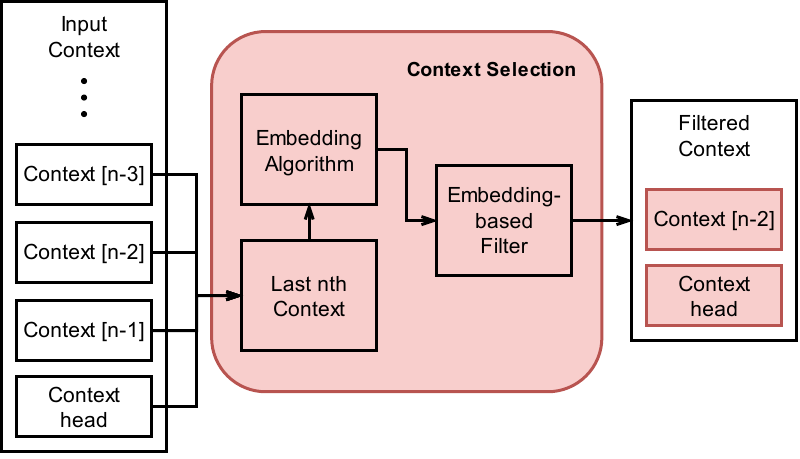}
\caption{Example schema for filtering a pipeline's input context.}
\end{figure}

In our method, each segment of the input context is processed through embedding models that capture semantic similarity and relevance. These embeddings are then matched against the task requirements or user query to evaluate their pertinence. By continuously applying this filtering, we ensure that only the most relevant pieces of information are retained. This helps maintain clarity and coherence in the generated outputs by reducing noise from unrelated context segments.

The use of embedding-based filters adds an adaptive layer capable of dynamically prioritizing context based on its relevance score. This process benefits from advancements in contextual embeddings that can factor in subtle semantic relationships, reducing the likelihood of context misalignment. Furthermore, by systematically pruning irrelevant data, the computational cost and memory usage during inference are also optimized, enhancing the system’s efficiency.

\section{Experiments}

\subsection{Synthetic Context-Aware System Dataset}
To benchmark our proposed techniques, we created a synthetic datasets mimicking natural prompt utterances and complete assistant conversations for a context-aware system. The synthetic data is generated using 135 of the most commonly searched topics from Google in 2024 over a broad range of domains. 

\begin{figure}[h]
\centering
\includegraphics[width=0.45\textwidth]{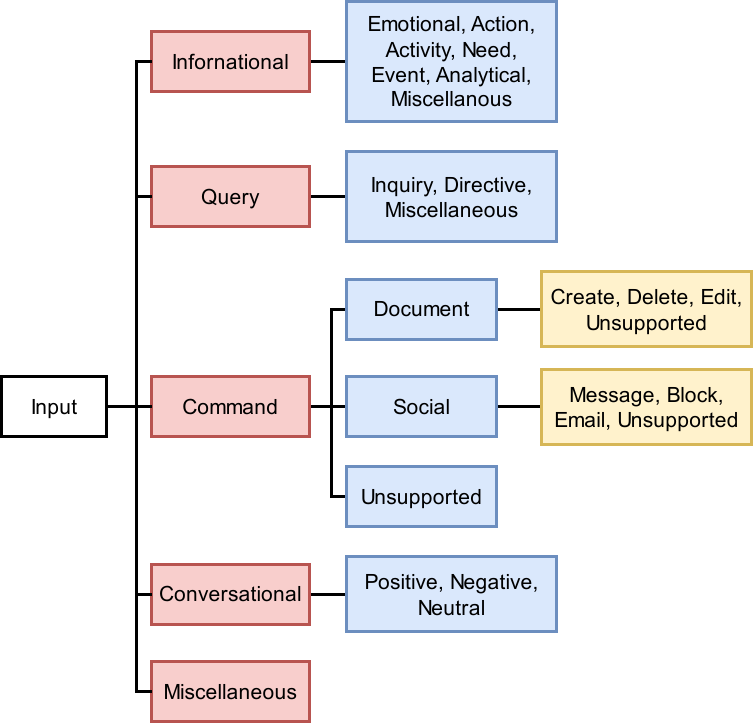}
\caption{Complete Context-Aware System Hierarchy, with the color indicating the schema node depth. Leaf nodes are aggregated for brevity.}
\label{fig:complete_hierarchy}
\end{figure}

First, we created a sample hierarchal schema for a hypothetical context-aware system shown in Fig. \ref{fig:complete_hierarchy}. This is to have a baseline guide on how to generate our data in a structured manner. Based on the hierarchal schema we created, we synthesized two datasets, namely Synthetic Single Prompt Dataset (SynPrompt), and Synthetic Assistant Conversations Dataset (SynAsst). \textbf{SynPrompt} contains synthesized utterances of multiple sentence lengths ranging from simple one sentence prompts up to complex five sentence paragraphs. Meanwhile \textbf{SynAsst} contains complete context trees of length ten and twenty (plus 1 synthetic user reference prompt).

One thing to note about \textbf{SynAsst} is that the conversation was generated with a percent probability of letting an a synthetic "user" derail the conversation by diverging from the main topic into another random one. In our case, we set the probability to diverge from the conversation as \(0.5\) or \(50\%\).

\subsection{Setup and Scoring}
LLM-based benchmarks often do not have well-defined baselines due to how unstructured and varied the domain of natural language is \cite{liEmbodiedAgentInterface2025}, so we propose a novel form of scoring to compute the performance of our techniques. We base our scoring on how consistent the LLM-agents generate a structured decision called \textit{Exponential Consistency Index} (ECI) with the given formula:

\begin{equation}
    \bar{S} =\frac{1}{d} \sum_{i=1}^{d} \left(\frac{\hat{k_i}}{k}\right)^\alpha
    \label{eq:eci}
\end{equation}
where \(\bar{S}\) is the consistency index of a given output to evaluate, \(d\) is the total number of items to evaluate for a given output entry, \(\hat{k_i}\) is the correct number of evaluations for a given item, \(k\) is the total amount of evaluations, and \(\alpha\) is the exponential penalty term that decides how much to penalize mistakes across \(k\) evaluations for an item.

\begin{figure}[h]
\centering
\includegraphics[width=0.45\textwidth]{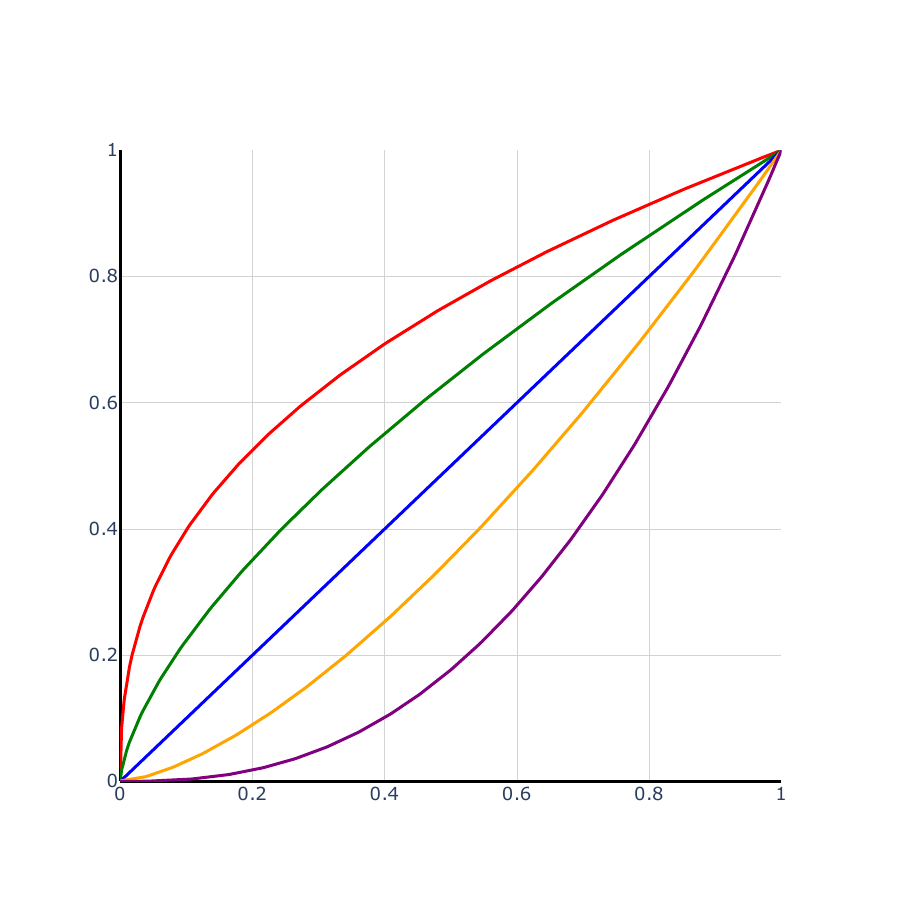}
\caption{Plot of the \textit{Exponential Consistency Index} given various \(\alpha\) values \(0.4, 0.65, 1.0, 1.5, 2.5\), with the \textit{x-axis} corresponding to the ratio of correct evaluations \(\hat{k_i}/{k}\). Higher \(\alpha\) values penalize slight evaluation mistakes more heavily.}
\label{fig:eci}
\end{figure}

For \textit{Semantic Decomposition}, the task is to evaluate the list of types of a given input sentence in a hierarchal fashion using the schema in Fig. \ref{fig:complete_hierarchy}.

For \textit{Selective Context Filtering}, the task is to select which pieces of context in the conversation tree is relevant to the latest \textit{"user"} prompt. We use both LLM-based evaluations and vector-based embeddings (threshold) to select the most relevant pieces of context. Furthermore, we calculate the accuracy of the extracted pieces of context by comparing it to whether the synthetic \textit{"user"} prompt was instructed to diverge from the main topic.

For all setups, the input prompt's context information will either be pre-processed or not. If the context information is decomposed beforehand, it will be injected at every step of the evaluation. This is to test whether a more contextualized generation will result to more consistent outputs.  

Alongside our datasets \textbf{SynPrompt} and \textbf{SynAsst}, we evaluate our methods over the OpenAssistant Conversations dataset (OASST1 and OASST2) \cite{kopfOpenAssistantConversationsDemocratizing2023} in order to evaluated our methods over natural prompts and conversations. 

\subsection{Results and Discussion}
For all the experiments, a subset of 150 entries will be utilized as the test datasets, in line with current standards for LLM-based evaluation setups \cite{gaoRARRResearchingRevising2023}. Furthermore, a value of \(\alpha = 2.5\) for ECI-based scoring will be used. We denote experiments without decomposing the initial context prompt as \textit{Normal}, and \textit{C.D.} if it was decomposed preemptively. For the LLM-agents, \textit{gpt-4o} and \textit{gpt-4o-mini} are chosen. 

In evaluating Semantic Decomposition, we perform ECI-based scoring over \textbf{SynPrompt} (1-5 sentences), \textbf{OASST1} and \textbf{OASST2}. We calculate the consistency of the decomposed schema list over \(k=5\) evaluations per entry.

\begin{table}[h]
    \centering
    \begin{tabular}{lSSSSSS}
        \toprule
        \multirow{2}{*}{Dataset} &
        \multicolumn{2}{c}{gpt-4o-mini} &
        \multicolumn{2}{c}{gpt-4o}  \\
        & {Normal} & {C.D.} & {Normal} & {C.D.} \\
        \midrule
        S1 & 0.860 & \textbf{0.862} & 0.821 & 0.863  \\
        S2 & 0.875 & \textbf{0.879} & 0.832 & 0.834  \\
        S3 & \textbf{0.914} & 0.913 & 0.857 & 0.886  \\
        S4 & \textbf{0.909} & 0.874 & 0.816 & 0.852  \\
        S5 & \textbf{0.912} & 0.857 & 0.788 & 0.844  \\
        OASST1 & \textbf{0.848} & 0.820 & 0.739 & 0.799  \\
        OASST2 & \textbf{0.830} & 0.817 & 0.735 & 0.825  \\
        \bottomrule
    \end{tabular}
    \caption{Semantic Decomposition mean ECI scores}
\end{table}

In the \textit{Selective Context Filtering} experiments, we perform ECI-based scoring over \textbf{SynAsst} (5-pair and 10-pair), along with \textbf{OASST1}. We calculate the consistency of the indices list of extracted context pieces over \(k=5\) evaluations per entry.

\begin{table}[h]
    \centering
    \begin{tabular}{lSSSS}
    \toprule
    \multirow{2}{*}{Dataset} &
    \multicolumn{2}{c}{gpt-4o-mini} &
    \multicolumn{2}{c}{gpt-4o}  \\
    & {Normal} & {C.D.} & {Normal} & {C.D.} \\
    \midrule
    10-Pair & \textbf{0.902} & 0.880 & 0.848 & 0.859 \\
    5-Pair  & \textbf{0.887} & 0.871 & 0.821 & 0.794 \\
    OASST1  & \textbf{0.826} & 0.754 & 0.683 & 0.590 \\
    \bottomrule
    \end{tabular}
    \caption{Selective Context Filtering mean ECI scores}
\end{table}

For calculate the accuracy of the extracted context list, we perform the evaluations only over \textbf{SynAsst} (5-pair and 10-pair). We utilize both LLM-based and embedding vector-based methods. The accuracy for LLM-based methods is calculated \(k=5\) evaluations per entry, while for vector-based methods, \(k=1\) only (because vector-embedding models produce consistent outputs).

\begin{table}[h]
    \centering
    \begin{tabular}{lSSSS}
        \toprule
        \multirow{2}{*}{Dataset} & \multicolumn{2}{c}{LLM Normal} & \multicolumn{2}{c}{LLM C.D.} \\
        \cmidrule(lr){2-3} \cmidrule(lr){4-5}
        & {gpt-4o-mini} & {gpt-4o} & {gpt-4o-mini} & {gpt-4o} \\
        \midrule
        5-Pair & 0.631 & 0.623 & 0.616 & 0.620 \\
        10-Pair & 0.602 & 0.604 & 0.611 & 0.605 \\
        \midrule
        \multicolumn{5}{c}{Vector-Based} \\
        \cmidrule(lr){2-5}
        & \multicolumn{2}{c}{text-embedding-3-small} & \multicolumn{2}{c}{text-embedding-3-large} \\
        \cmidrule(lr){2-3} \cmidrule(lr){4-5}
        5-Pair & \multicolumn{2}{c}{\textbf{0.700}} & \multicolumn{2}{c}{0.649} \\
        10-Pair & \multicolumn{2}{c}{\textbf{0.754}} & \multicolumn{2}{c}{0.716} \\
        \bottomrule
    \end{tabular}
    \caption{Selective Context Filtering mean accuracy}
\end{table}

In analyzing the results of the experiments, we can see that having no pre-decomposed context being injected in all stages of evaluation produces more consistent results. We hypothesize this is because having too context tokens during evaluation can derail some types of evaluation due to ambiguity. 

However, too little context information seems to lessen the consistency, with the 3-sentence subset of \textbf{SynPrompt} with no Context Decomposition having the highest Semantic Decomposition ECI scores for both \textbf{Normal} and \textbf{C.D.} We posit that a balance needs to be achieved in how much context tokens needs to be passed on during all stages of evaluation. 

For Selective Context Filtering accuracy, the vector-based methods performed marginally better than LLM-based ones. We initially hypothesized that LLM-based methods would perform better due to having more complex internal mechanisms in transformer models, but it seems that embedding vector-based methods still works better for similarity and relevancy calculations between different pieces of text.

Surprisingly, \textit{gpt-4o-mini} produces more consistent evaluations across the board despite being a "weaker" model than \textit{gpt-4o} due to having a lower model parameter count. This is the case for Selective Context Filtering accuracy as well, with \textit{text-embedding-3-small} having the most accurate relevancy extraction. We currently do not know the reasoning behind why the smaller models perform better on our techniques, so more future work still needs to be done with regards to analysis in this aspect.

\section{Conclusion}

In conclusion, the integration of advanced techniques such as Semantic Decomposition and Selective Context Filtering within context-aware systems significantly enhances the capability and adaptability of LLM-based pipelines. These methodologies provide a structured framework that improves information parsing, ensures contextual relevance, and facilitates dynamic interactions between systems and users. By implementing these strategies, systems can deliver more contextually coherent and user-tailored responses, proving beneficial across diverse fields such as customer service, healthcare, education, and finance. Leveraging these enhanced capabilities will be vital in addressing complex challenges, optimizing automated workflows, and enabling more sophisticated applications. Ultimately, this approach underscores the potential of LLMs to transform how systems interpret and respond to contextual information, paving the way for more intelligent and responsive technology solutions. Future work will aim at refining and evaluating these techniques further, exploring additional decomposition techniques, utilizing other RAG algorithms, and testing on a wider set of models. 

\bibliographystyle{ACM-Reference-Format}
\bibliography{references}

\end{document}